# The analogy theorem in Hoare logic

Nikitin Nikita[1,2]

[1]College of New Materials and Nanotechnologies. National University of Science & Technology (MISIS), Leninskii prosp, 4, 119049, Moscow, Russia;

[2]Spark Plasma Sintering Research Laboratory, Moscow State University of Technology "STANKIN", Vadkovsky per. 1, Moscow, 127055, Russia

Corresponding authors: nikitin5@yandex.ru (Nikitin Nikita)

## Abstract

The introduction of machine learning methods has led to significant advances in automation, optimization, and discoveries in various fields of science and technology. However, their widespread application faces a fundamental limitation: the transfer of models between data domains generally lacks a rigorous mathematical justification. The key problem is the lack of formal criteria to guarantee that a model trained on one type of data will retain its properties on another.

This paper proposes a solution to this problem by formalizing the concept of "analogy" between data sets and models using first-order logic and Hoare logic. We formulate and rigorously prove a theorem that sets out the necessary and sufficient conditions for analogy in the task of knowledge transfer between machine learning models.

Practical verification of the analogy theorem on model data obtained using the Monte Carlo method, as well as on MNIST and USPS data, allows us to achieving F1-scores of 0.84 and 0.88 for convolutional neural networks and random forests, respectively.

The proposed approach not only allows us to justify the correctness of transfer between domains but also provides tools for comparing the applicability of models to different types of data.

The main contribution of the work is a rigorous formalization of analogy at the level of program logic, providing verifiable guarantees of the correctness of knowledge transfer, which opens new opportunities for both theoretical research and the practical use of machine learning models in previously inaccessible areas.

## 1. Introduction

Over the course of mathematical logic's existence, two main approaches to constructing logical arguments have been developed in detail, namely deductive and inductive. The third approach, "analogical," remains at the level of intuitive knowledge and does not have a strict formulation in mathematical logic.

However, attempts to construct rigorous mathematical theorems with a meaning close to the concept of "analogy" have been made by several researchers. For example, Craig-Lindon's theorem on interpolation [1-3] allows us to establish a relationship between different mathematical theories in first-order logic and may be valid when modified for other logics. The Craig-Lindon theorem on interpolation is formulated as [1]:

«*Let S and T be sentences of language L such that S => T. Then there exists a sentence S0 of this language L such that S => $S^0$, $S^0$ => T, and that the relation symbol occurs positively in $S^0$ only if it occurs positively in both S and T, and negatively in $S^0$ only if it occurs negatively in both S and T.*»

In its original formulation, Craig-Lindon's theorem only considers logical transition and does not consider "analogy" as such, which limits its scope to only the area where strict logical transition exists.

The second approach, which can be attributed to the concept of "analogy," relates to interpretation and conservative extension in model theory. The main concept in this case can be represented as [4,5]:

*"If theory $T_1$ can be interpreted in $T_2$, then the truth of statement $T_1$ is transferred to $T_2$."*

This formulation implies the complete interpretation of one theory in another, which is not required in the case of analogy.

The next approach to constructing logical conclusions by analogy may be abduction (a method of reasoning created by Charlie Peirce) [6-8]. For example, we can express our reasoning as follows:

*"Let B be observed and it be known that A→B, then it is assumed that A is the cause of B."* However, in this case, we are not talking about analogy, but only about the formulation of a plausible hypothesis, which limits the application of these approaches for comparing a broad class of objects and laws.

One can attempt to formulate a theorem of analogy in category theory approaches, through functors and natural transformations [9,10]. Let us assume the following:

*"Let functor F:C→D transfer structures from one category to another, then if F preserves properties, objects C and D behave 'analogously'."* In this case, the forgetful functor from Grp to Set "forgets" the group structure. However, with this approach, categories will operate at the level of morphisms rather than at the level of logical formulas, which limits the application of the theorem.

Another approach to solving the analogy problem may be the structural mapping proposed by Gentner [11] and can be represented as analogies of the form:

| | |
|---|---|
| $A: B :: C: D$ | (1) |

The problems of structural mapping that limit its practical application are the complexity of scaling for real knowledge and the ability to operate only with correctly specified structures [12].

As an alternative to the above approaches, learning transfer models [13-15] used in machine learning methods can be considered. Currently, the task of learning transfer is one of the most pressing issues. This is because models built using machine learning methods are limited to a specific data set belonging to a specific type of object. Applying the model to unknown data is difficult or impossible in principle.

Summarizing all the above, we can conclude that despite numerous approaches, strict formalization of the concept of “analogy” at the level of logical formulas, applicable to the task of knowledge transfer between models and models obtained by machine learning methods, remains underdeveloped.

One possible solution to this problem is to use Hoare logic to formalize knowledge transfer methods. The introduction of Hoare logic into the context of machine learning is driven by the need for strict guarantees of model correctness when transferring between domains. This logic allows us to formally describe that, under certain conditions imposed on the input data, the properties of the model (e.g., accuracy, stability) will be preserved on new data. This approach is fundamentally different from the empirical methods that dominate modern ML and paves the way for automated verification of model reliability, which is critical for their implementation in responsible applications [16-19].

Hoare's classical logic allows us to set preconditions and postconditions for each stage of data processing or training, thereby providing strict guarantees of the correctness of transformations—for example, that input data of a certain distribution leads to output with specified characteristics, or that the procedure for transferring a model between domains preserves accuracy invariants. This approach not only increases trust in ML systems but also paves the way for automated verification of their properties using modern static analysis tools and formal methods. This is especially valuable in tasks where the safety, reliability, and explainability of artificial intelligence must be ensured, which is becoming the standard for modern industry and science [16-19].

The goal of our work is to formulate, prove, and demonstrate the applicability of the analogy theorem in Hoare logic, which sets formal conditions for the correctness of knowledge transfer from one machine learning model to another.

## 2. Materials and methods

### 2.1. Formulation of the analogy theorem in predicate logic

Since statements in Hoare logic (preconditions, postconditions, invariants) are expressed in first-order logic (FOL), the theorem is first considered within the basic calculus—first-order logic. This guarantees its mathematical rigor and universality: the result is applicable not only to the verification of classical imperative programs, but also to the analysis of data transformations and properties of machine learning models, if they can be correctly described in terms of FOL. Such a proven theorem can be transferred to Hoare logic through standard interpretation rules, preserving its proven truth and applicability to software and ML systems [16, 20, 21].

To formulate and prove the analogy theorem in this paper, we use the language of first-order predicate logic with the equality symbol introduced. Accordingly, we operate with the language with the signature Ω=⟨Str,Cnst,Fn,Pr⟩, where Str is a non-empty set whose elements will be called object types; Cnst is a set of constants of the language Ω; Fn is a set whose elements will be called functional symbols of the language Ω; and Pr is a non-empty set whose elements will be called predicate symbols of the language Ω [22-25]. We will use the names of variables x and y as subject variables. We will consider logical connectives and quantifiers to be symbols of the predicate logic language:

- ∧ - conjunction (logical "and");
- ∨ - disjunction (logical "or");
- → - implication, "if..., then," "entails";
- ¬ - negation (logical "not");
- ↔ - equivalence ("if and only if…");

Logical quantifiers are represented as:

- ∀ - universality (generalization), in natural language "for everyone";
- ∃ - existence, "exists".

Before considering the relationship between theories, laws, formulas, and variables, let us define a theory in predicate logic. Accordingly, we will assign certain predicate symbols F and L to each variable x and y, which are applied to variable s, forming formulas *F(s)* and *L(s)*, interpreted in theories $T_C$ and $T_M$, where the theories $T_C$ and $T_M$ are consistent Con($T_C$)∧Con($T_M$), and x and y are variables of type s [26, 27].

Then, let there be C and M, which are closed formulas (laws) such that:

- $T_C \vdash C$, i.e., there is a theory that generalizes *F(s)*,
- $T_M \vdash M$, i.e., there is a theory that generalizes *L(s)*.

Let us formulate the relationship between laws C and M and formulas F(s) and L(s) as axioms, namely:

| | |
|---|---|
| $\forall x(F(x) \to C)$ | (A1) |
| $\forall x(L(x) \to M)$ | (A2) |

Let us assume that there is a hypothesis $H_1(s)$ such that $F(s)$ and $L(s)$ behave "identically" with respect to $C$ and $M$. And in the case of hypothesis $H_2(s)$, such that $F(s)$ and $L(s)$ behave differently with respect to $M$, i.e., the analogy is partially violated. Then we will say that $H_1$ establishes an analogy between $F(s)$ and $L(s)$ if: $\forall s\ (H_1\ (s)(F(s)\leftrightarrow L(s)))$. Hypothesis $H_2$ breaks the analogy if: $\exists s(H_2(s)\wedge F(s)\wedge\neg L(s))$. We introduce the functional symbol $D{:}s\times s\rightarrow R$, where $R$ is the set of real numbers and the predicate symbols ≤, > on $R$, and the hypotheses $H_1(x)$ and $H_2(x)$ are defined through them as: $H_1(x)\equiv(D(x,x_0)\leq\varepsilon)$ and $H_2(x)\equiv(D(x,x_0)>\varepsilon)$.

For the functional symbol $D{:}s\times s\rightarrow R$ and the constant ε, we introduce a set of axioms that define its properties [28, 29]:

| | | |
|---|---|---|
| $\forall x(D(x,x)=0)$ | Reflexivity | (A3) |
| $\forall x\forall y\ (D(x,y)\geq\ 0\ )$ | No negativity | (A4) |
| $\forall x\forall y\big(D(x,y)=0\ \leftrightarrow(x=y)\big)$ | Axiom of identity | (A5) |
| $\forall x\forall y(D(x,y)=D(y,x))$ | Symmetry | (A6) |
| $\forall x\forall y\forall z(D(x,z)\leq D(x,y)+D(y,z))$ | Inequality of a triangle | (A7) |
| $\forall x\forall y(D(x,y)\leq\delta\rightarrow(F(x)\leftrightarrow F(y))$ | Stability of property F | (A8) |
| $\forall x\forall y(D(x,y)\leq\delta\rightarrow(L(x)\leftrightarrow L(y))$ | L property stability | (A9) |
| $F(x_0)\leftrightarrow L(x_0)$ | Reference similarity | (A10) |
| $\exists x\Big(D(x,x_0)>\delta \wedge\Big(\big(F(x)\wedge\neg L(x)\big)\vee\big(\neg F(x)\wedge L(x)\big)\Big)\Big)$ | Non-triviality outside δ | (A11) |

For stability axioms A8 and A9 to provide sufficient conditions, it is necessary that functions F and L be regular. This requirement can be formulated as a condition on the regularity of predicates F and L.

**Regularity condition.**

Let predicates F and L be defined by threshold functions $f_F{:}\,s\longrightarrow\mathbb{R}, f_L{:}\,s\rightarrow\mathbb{R}$, and let there be constants $\tau>0, L_F>0, L_L>0$ such that:

1. $\forall x(F(x)\leftrightarrow f_{F(x)}\geq 0)$;
2. $\forall x(L(x)\leftrightarrow f_{L(x)}\geq 0)$;
3. $|f_{F(x)}|\geq\tau$ и $|f_{L(x)}|\geq\tau$, for all x such that $D(x,x_0)\leq\delta$;
4. $|f_{F(x)}-f_{F(y)}|\leq L_F\cdot D(x,y)$ for everyone $x,y\in s$
5. $|f_{L(x)}-f_{L(y)}|\leq L_L\cdot D(x,y)$ for everyone $x,y\in s$

6. $\delta \le \min\left(\frac{\tau}{L_F}, \frac{\tau}{L_L}\right)$

That is, when formulating and proving the theorem, we will assume that the predicates F and L are defined through threshold functions $f_F$ and $f_L$, which have Lipschitz properties in the metric D. At the same time, $f_F$ and $f_L$ are external interpretations, while the representation of predicates F and L is preserved within the language.

In first-order predicate logic, the equality symbol = is used as a logical constant for all types, with the usual axioms of reflexivity, symmetry, transitivity, and substitution.

Based on the above, we formulate the axioms defining the analogy hypothesis as follows:

| | | |
|---|---|---|
| Completeness of hypotheses | $\Gamma \vdash \forall x (H_1(x) \vee H_2(x))$ | (M1) |
| Mutual exclusion of hypotheses | $\Gamma \vdash \neg \exists x (H_1(x) \wedge H_2(x))$ | (M2) |
| Connection with metrics D(s) | $H_1(x) \equiv (D(x, x_0) \le \varepsilon)$<br>$H_2(x) \equiv (D(x, x_0) > \varepsilon)$<br>*Where $x_0$ is the reference element; ε is a constant that defines the boundaries of the hypothesis, and D is a metric of similarity between hypotheses.* | (M3) |

Summarizing the above assumptions, we formalize the task of knowledge transfer between machine learning models as a problem of transferring properties between data domains. Then, let the signature be given:

| | |
|---|---|
| $\Omega = \langle \{s, \mathbb{R}\}, \{\varepsilon: \mathbb{R}, \delta: \mathbb{R}, 0: \mathbb{R}, x_0: s\}, \{D: s \times s \to \mathbb{R}\}, \{F, L, \le: \mathbb{R} \times \mathbb{R}, >: \mathbb{R} \times \mathbb{R}\} \rangle$ | (2) |

Where s is an arbitrary non-empty set (data type); R – when formulating the analogy theorem and its proof, we will consider it as a standard structure of real numbers with field and order axioms. Then axioms A3-A7 and A10 will define the metric in the usual analytical sense; $x_0$ is the reference model; ε is the boundary of analogy fulfillment; δ is the stability boundary; 0 is zero; D is the metric on s; F, L are predicates (model properties); ≤,> are standard order relations.

**Analogy theorem:**

Let *ε<δ, Γ={A3-A11},* where *{A3-A11}* are metric axioms and hypotheses *M1-M3*, and functions F and L satisfy the regularity condition, then the following statement is true:

| | |
|---|---|
| $\Gamma \vdash \forall x [(D(x, x_0) \le \varepsilon) \to (F(x) \leftrightarrow L(x))]$ | (3) |

Such as:

| | |
|---|---|
| $\Gamma \vdash \exists x[(D(x, x_0) > \varepsilon) \wedge F(x) \wedge \neg L(x)]$ | (4) |

**Evidence for the analogy theorem:**

Suppose that $\Gamma \vdash D(x, x_0) \leq \varepsilon$ and, in accordance with the conditions of the theorem $\Gamma \vdash \varepsilon < \delta$, in the standard linear order $\mathbb{R}$ and based on the properties of transitivity, we obtain that $\Gamma \vdash D(x, x_0) \leq \delta$. Accordingly, if $y=x_0$, then using axiom A8, we obtain $\Gamma \vdash \Big(D(x, x_0) \leq \delta \rightarrow (F(x) \leftrightarrow F(x_0))\Big)$, and applying the Modus Ponens rule, we have:

| | |
|---|---|
| $\dfrac{D(x, x_0) \leq \delta, D(x, x_0) \leq \delta \rightarrow (F(x) \leftrightarrow F(x_0))}{F(x) \leftrightarrow F(x_0)}$ | (5) |

That is, $\Gamma \vdash F(x) \leftrightarrow F(x_0)$. From axiom A11, we have $\Gamma \vdash F(x_0) \leftrightarrow L(x_0)$, and given that $y=x_0$ and axiom A9, we have $\Gamma \vdash \Big(D(x, x_0) \leq \delta \rightarrow (L(x) \leftrightarrow L(x_0))\Big)$. Using the Modus Ponens rule, we have:

| | |
|---|---|
| $\dfrac{D(x, x_0) \leq \delta, D(x, x_0) \leq \delta \rightarrow (L(x) \leftrightarrow L(x_0))}{L(x) \leftrightarrow L(x_0)}$ | (6) |

That is, $\Gamma \vdash L(x) \leftrightarrow L(x_0)$, using the properties of transitivity, as well as $\Gamma \vdash F(x) \leftrightarrow F(x_0)$, $\Gamma \vdash F(x_0) \leftrightarrow L(x_0)$, and $\Gamma \vdash L(x) \leftrightarrow L(x_0)$, we obtain that $\Gamma \vdash F(x) \leftrightarrow L(x)$. Generally assessing the obtained result for all x, we obtain statement (2) of the analogy theorem:

| | |
|---|---|
| $\Gamma \vdash \forall x[(D(x, x_0) \leq \varepsilon) \rightarrow (F(x) \leftrightarrow L(x))]$ | (7) |

Which is what we wanted to prove.

Now let's consider the proof of statement (3). It follows directly from axiom (A11) that there exists a point for which $D(x, x_0) > \delta$ and the analogy between F and L is violated. In accordance with the condition of the theorem ε<δ, it follows that for this point the statement $D(x, x_0) > \varepsilon$ is also true. Further, in accordance with axiom (A11):

| | |
|---|---|
| $\exists x \Big(D(x, x_0) > \delta \wedge \big((F(x) \wedge \neg L(x)) \vee (\neg F(x) \wedge L(x))\big)\Big)$ | (8) |

Then we can consider the case where there exists x' such that $F(x') \wedge \neg L(x')$. (The case $\neg F(x') \wedge L(x')$ can be proven symmetrically). From the condition of the theorem ε<δ, it follows from $D(x', x_0) > \delta$ that $D(x', x_0) > \varepsilon$. Thus, for x', the following statement is true:

| | |
|---|---|
| $D(x', x_0) > \varepsilon \wedge F(x') \wedge \neg L(x')$ | (9) |

Using the rule for introducing the quantifier ∃, we obtain the possibility of considering two options $(F(x') \wedge \neg L(x'))$ and $(\neg F(x') \wedge L(x'))$, whereby the choice of either option does not lead to a loss of generality, and taking into account (9), we obtain:

| | |
|---|---|
| $\Gamma \vdash \exists x\big(D(x, x_0) > \varepsilon \land F(x) \land \neg L(x)\big)$ | (10) |

Thus, we obtain statement (4) of the analogy theorem, which was to be proven.

The presented analogy theorem in first-order logic can be interpreted in machine learning and Hoare logic as follows: when performing axioms *Г*, the transfer of knowledge from one machine learning model to another is guaranteed within the ε-neighborhood of the reference domain. Accordingly, if the transfer conditions are not met, the result is not guaranteed [16, 30-32]. To ensure a wider application of the analogy theorem, we will present its formulation directly in Hoare logic. In doing so, we will consider that *F(x)* and *L(x)* can be interpreted as program invariants or data properties in machine learning, while the ε-neighborhood acts as the domain of correct transfer of the property or invariant.

## 2.2. Formulation of the analogy theorem in Hoare logic

To apply the analogy theorem to data transfer and algorithm comparison, let us introduce a formal description within the framework of Hoare logic [16]. Let *S* be the set of program states, $s_0 \in S$ be the reference state, $F, L{:}S \to \{true, false\}$ be predicates describing the properties of the states of the source and target models, $D{:}S \times S \to R \geq 0$ be a metric (satisfying the standard axioms A3-A7 of the previous part) [29], ε>0 is the analogy boundary; $\delta > 0$ is the stability boundary; $\gamma \geq 0$ is the execution stability parameter. In Hoare logic, statements about program behavior are expressed as triples *{P}S{Q}*, where *P* is the precondition, *S* is the program, and *Q* is the postcondition. In our case, considering program *S* as a state transformer, we are interested in the correctness of predicates *F* and *L* before and after the execution of *S*.

Having described Hoare's logic language, we can formulate the main conditions of the analogy theorem. Let us consider the first condition of the theorem, namely the condition of local equivalence in the ε neighborhood. So, let us assume that for all states s within a radius *ε* of $s_0$, the execution of the program preserves the equivalence of F and L. The second condition of the theorem will be the condition of non-triviality or the existence of a violation outside the ε-neighborhood. That is, there exists a state *s* outside the ε-neighborhood for which, after the execution of the program, *F* is true, and *L* is false. And the last, third property is the local property of F with parameter δ. We will assume that if states $s_1$ and $s_2$ are close (within radius δ), then after executing the program, F behaves identically. Based on the above for program S, metric D, and predicates F and L, the data model and predicates are [16,33,34]:

| | |
|---|---|
| $\forall s\left(D(s, s_0) \leq \varepsilon \to \big(F(s) \leftrightarrow L(s)\big)\right)$ | (U1) |
| $\exists s\big(D(s, s_0) > \varepsilon \land F(s) \land \neg L(s)\big)$ | (U2) |
| $\forall s_1 \forall s_2 \left(D(s_1, s_2) < \delta \rightarrow \left(F(s_1) \leftrightarrow F(s_2)\right)\right)$ | (U3) |

Where $\gamma \geq 0$ is the stability parameter; $\varphi_S: S \rightarrow S$ is the state transformation by program *S*. Program *S* has the following properties:

| | |
|---|---|
| γ- stable: $\forall s \ (D(\varphi_S(s), s) \leq \gamma)$ | (C1) |
| S preserves properties F: $\forall s \left(F(s) \rightarrow F\left(\varphi_S(s)\right)\right)$ | (C2) |

Having defined all the conditions, we will formulate the analogy theorem in Hoare logic.

**Analogy theorem:**

Let program S have properties C1 and C2. Then:

| | |
|---|---|
| $\{D(s, s_0) \leq \varepsilon - \gamma\}S\{F\left(\varphi_S(s)\right) \leftrightarrow L\left(\varphi_S(s)\right)\}$ | (U4) |
| $\{D(s^*, s_0) > \varepsilon + \gamma \wedge F(s^*)\}S\{\neg L\left(\varphi_S(s^*)\right)\}$ | (U5) |
| $\{D(s_1, s_2) < \delta - 2\gamma\}S\{F\left(\varphi_S(s_1)\right) \leftrightarrow F\left(\varphi_S(s_2)\right)\}$ | (U6) |

**Proof of the analogy theorem in Hoare logic:**

From C1, it follows that for $\forall s, D(\varphi_s(s), s_0) \leq \gamma$, and from precondition U4, that $D(s, s_0) \leq \varepsilon - \gamma$, then by the triangle inequality we obtain:

| | |
|---|---|
| $D(\varphi_s(s), s_0) \leq D(\varphi_s(s), s) + D(s, s_0) \leq \gamma + (\varepsilon - \gamma) = \varepsilon$ | (11) |

Then, from condition U1, we obtain: $D(\varphi_s(s), s_0) \leq \varepsilon \Rightarrow \left[F\left(\varphi_s(s)\right) \leftrightarrow L\left(\varphi_s(s)\right)\right]$, now formalized in Hoare logic as:

| | |
|---|---|
| $P' \rightarrow P, \{P\}S\{Q\}, Q \rightarrow Q'/\{P'\}S\{Q'\}$ | (12) |

Where $P' = D(s, s_0) \leq \varepsilon - \gamma, Q' = F\left(\varphi_s(s)\right) \leftrightarrow L\left(\varphi_s(s)\right)$. Which is what we wanted to prove. Now let's look at the proof of condition 5. From condition (U2) it follows that $\exists s^*, [D(s^*, s_0) > \varepsilon \wedge F(s^*) \wedge \neg L(s^*)]$, and from C1, that $\varphi_s(s^*)$ is defined and $D(\varphi_s(s^*), s^*) \leq \gamma$, then the key observation will be $D(\varphi_s(s^*), s_0) \geq D(s^*, s_0) - D(\varphi_s(s^*), s^*) > \varepsilon - \gamma$, but for direct application, ε is required, not *ε-γ*. In this case, we will strengthen condition C2 to ensure the stability of the disturbance, as follows:

| | |
|---|---|
| $\exists s^*, \forall \varphi_s, \left[D(s^*, s_0) > \varepsilon + \gamma \Rightarrow \neg L\left(\varphi_s(s^*)\right)\right]$ | (U2.1) |

Then, when $D(s^*, s_0) > \varepsilon + \gamma, F(s^*)$ is preserved, subject to condition C2, and it will be guaranteed that $\neg L\left(\varphi_s(s^*)\right)$. Consequently, the correct Hoare triplet in this case will be as follows:

| | |
|---|---|
| $\{D(s^*, s_0) > \varepsilon + \gamma \wedge F(s^*)\}S\{\neg L\left(\varphi_s(s^*)\right)\}$ | (13) |

Which is what we need to prove. Let us consider the proof of the last third condition in the analogy theorem formulated in Hoare's logic. There is condition (U3) $D(s_1, s_2) < \delta - 2\gamma$ and stability condition U1:

| | |
|---|---|
| $[\forall s \exists \varphi_S: (post(S, s) = \varphi_s(S) \wedge D(\varphi_S(s), s) \leq \gamma)]$ | (U1.1) |

Applying U1.1 sequentially to $s_1$ and $s_2$, we obtain $D(\varphi_S(s_1), s_1) < \gamma, D(\varphi_S(s_2), s_2) < \gamma$ and, by the triangle inequality, we obtain:

| | |
|---|---|
| $D(\varphi_S(s_1), \varphi_s(s_2)) \leq D(\varphi_S(s_1)) + D(s_1, s_2) + D(s_2, \varphi_S(s_2))$ | (14) |

Substituting the estimates, we obtain:

| | |
|---|---|
| $D(\varphi_S(s_1), \varphi_S(s_2)) \leq \gamma + (\delta - 2\gamma) + \gamma = \delta$ | (15) |

Then, considering the applicability condition, we obtain:

| | |
|---|---|
| $D(\varphi_S(s_1), \varphi_S(s_2)) < \delta$ | (16) |

Applying condition U3 to the transformed states (16), we obtain:

| | |
|---|---|
| $D(\varphi_S(s_1), \varphi_S(s_2)) < \delta \Rightarrow [F(\varphi_S(s_1)) \leftrightarrow F(\varphi_S(s_2))]$ | (17) |

That is, condition 3 of Hoare's analogy theorem in logic. The formalized representation of the inference is based on the application of the consequence rule with an invariant:

| | |
|---|---|
| $P' \rightarrow P, \{P\}S\{Q\}, Q \rightarrow Q'/\{P'\}S\{Q'\}$ | (18) |

Where $P' = D(s_1, s_2) < \delta - 2\gamma$ and $Q' = F(\varphi_S(s_1)) \leftrightarrow F(\varphi_S(s_2))$, from which the resulting Hoare triple has the form:

| | |
|---|---|
| $\vdash [\{D(s_1, s_2) < \delta - 2\gamma\}S\{F(\varphi_S(s_1)) \leftrightarrow F(\varphi_S(s_2))\}]$ | (19) |

Which is what needed to be proven.

The presented analogy theorem has an interesting consequence when $\gamma < \min(\varepsilon, \frac{\delta}{2})$: all statements of the theorem are preserved with the following modified parameters: the ε-neighborhood is equal to $\varepsilon' = \varepsilon - \gamma$, and $\delta' = \delta - 2\gamma$, which can be interpreted as the stronger the transformation ↑γ, the narrower the region of analogy. The presented reasoning also implies restrictions on the model related to axiom M4, namely, if the transformations lead to $\gamma > \min(\varepsilon, \frac{\delta}{2})$, then the theorem ceases to work.

A comparative analysis of the analogy theorem in FOL and Hoare logic shows that in the case of determinism S, i.e., when S assigns exactly one state $s'$ to one state $s$, it can be asserted that Hoare's triple {P}S{Q} means that if the precondition P(s) is satisfied, then for the unique s' Q(s') is true. In this case, there is a direct translation from statement 3 of the theorem in FOL logic to statement 2 of the theorem in Hoare logic. In fact, this means that if the initial state is within δ-2γ, then after the operation of deterministic S, the state where the analogy is broken is guaranteed to be obtained. In the case of a non-deterministic S, Hoare's triple {P} S {Q} means that for all possible outputs $s' \in Post(S, s)$, i.e., if *P(s)*, then Q(s'). Whereas statement (3) of the analogy theorem asserts the existence of one bad element, not that all elements will be bad.

For the non-deterministic case U2 of the theorem, it should be represented as (in demonic semantics):

| | |
|---|---|
| $\{D(s, s_0) \leq \delta - 2\gamma\} S \{\exists s' \in Post(S, s): D(s', s_0) > \varepsilon \wedge F(s') \wedge \neg L(s')\}$ | (20) |

Then U3 in formulation 32 will refer to the existence of at least one element that does not correspond to the analogy, rather than all of them, and will correspond to statement 3 of the theorem on analogy in FOL logic. In the case of a weaker (angelic semantics) representation of statement 2, it will look like this:

| | |
|---|---|
| $\{D(s, s_0) \leq \delta - 2\gamma\} S \{D(s', s_0) > \varepsilon \wedge F(s') \wedge \neg L(s')\}$ | (21) |

Formulation 33 would mean that there is an execution that leads to a violation of the analogy but does not guarantee that all executions will be like that. From a practical point of view, it is necessary to determine the methods for calculating D, ε, δ, and γ. To do this, let us consider the existing methods for determining the distances between different data.

### 2.3. Options for evaluating the similarity metric D(s, $s_0$).

Let's consider the main approaches to defining similar metrics between models and data that currently exist. Table 1 lists metrics that are often used for numerical and real data, while Table 2 lists metrics for categorical, string, and time data. Each metric has its own area of application. For example, Euclidean distance measures the absolute difference between points and is widely used in clustering and regression analysis, while cosine similarity reflects the angular similarity of vectors and is particularly useful for text analysis and recommendation systems.

**Table 1.** List of similar metrics between numerical and real data

| Distance name | Brief description | Applicability in ML |
|---|---|---|
| Euclidean [35] | Minimum distance between points in n-dimensional space | Clustering, k-NN, regression [36,37] |
| Manhattan [36] | Sum of absolute differences in coordinates | Regression, feature selection, noise-sensitive methods [37,38] |
| Minkovsky [39] | Generalization of Euclidean and Manhattan distances, parameterized by p | Classification, clustering [39] |
| Chebyshev [40] | Maximum difference in coordinates | Detection of emissions [40] |
| Mahalanobis [41] | Considering the covariance structure of the features | Statistical classification, multidimensional data analysis [42] |
| Canberra [43] | Weighted sum of differences to the sum of values | Clustering, bioinformatics [43,44] |
| Bhattacharya [45] | Measures the similarity between probability distributions | Bayesian classification, image recognition |
| Hellinger [46] | A measure of distribution difference associated with Bhattacharya | Distribution analysis, bioinformatics |
| Correlation (Pearson, Spearman, Kendall) [47-49] | Linear and rank correlation measures | Feature selection, time series analysis |
| Cosine similarity [50] | The angle between vectors does not depend on length. | NLP, recommendation systems [50] |

Table 2 presents similar metrics applicable to data other than numerical and real data.

**Table 2.** Similarity metrics for data other than numerical and real data

| Data type | Metrics | Brief description | Applicability in ML |
|---|---|---|---|
| Categorical/ Sets | Jacquard index [51,52] | The proportion of overlapping elements to the union for sets or binary vectors. | Classification, recommendation systems, graph analysis [52] |
| Categorical/ Sets | Sørensen–Dice coefficient [53-55] | Like Jacquard, but with a different weight for crossing. | Classification, bioinformatics, image analysis [53-55] |
| Strings/character sequences | Hemming distance [56, 57] | The number of positions in which the values differ. | Coding, bioinformatics, digital signal recognition [57] |
| Strings/character sequences | Levenshtein distance (edit distance) [58,59] | The minimum number of insert, delete, and replace operations required to transform one string into another. | NLP, bioinformatics, error correction [59] |
| Strings/character sequences | Distance Li [60,61] | An edit distance variant that considers the positions of substrings. | NLP, text search, sequence analysis [60, 61] |
| Strings/character sequences | Yaro-Winkler distance [62,63] | Modified Yaro distance, considers common substrings and shifts. | Name comparison, analysis of personal data databases [62,63] |
| Time series | Dynamic Time Warping (DTW) [64, 65] | Flexible time series alignment that takes time shifts into account. | Time series analysis, signal processing, speech recognition [64,65] |
| Specialized metrics | Internal work (dot product) [66,67] | Scalar product of vectors. | Deep learning, feature space construction [66,67] |
| Specialized metrics | Motyka, Kulczynski, Bray–Curtis etc. [68, 69] | Can consider structure, semantics, tree | In recognition tasks, text, etc. [68] |

In addition to the metrics listed in Tables 1 and 2, there are similar metrics based on the evaluation of the distribution of random variables. To formalize the requirements for the similarity metric $D(s, s_0)$ within the framework of the analogy theorem, it is proposed to extend the axioms by including probability measures. For example, axiom M3 can be formulated in terms of the probability of analogy. This allows the use of mathematical statistics methods to estimate the differences between data distributions in different domains:

$$P(D(s,s_0) \leq \varepsilon) \geq 1 - \alpha \quad \text{(M3.1)}$$

Where P is the probability that the analogy holds.

For data described by probability distributions, statistical metrics and divergences such as Kulback–Leibler, Hellinger, Wasserstein, and others are used (see Table 3). These metrics allow us to compare not only individual points, but also entire distributions, which is especially important for tasks of transfer learning and analysis of shifts in data.

**Table 3.** List of statistical metrics and divergences used to evaluate D(s,s0).

| Metrics | Differences |
|---|---|
| Statistical distance [70] | Kulbak–Leibler divergence [76] |
| Helling distance [71] | Rényi divergence [77] |
| Levi-Prokhorov metric [72] | Jensen–Shannon divergence [78] |
| Wasserstein metric [73] | Ball deviation [79] |
| Mahalanobis distance [74] | Bhattacharya distance [80] |
| Integral probability metrics [75] | f-divergence [81] |
| | Discrimination index [82] |

Given the restrictions imposed on the metric D(s, s0) by axioms A3-A7 within the framework of the formulation of the analogy theorem in predicate logic, the metrics given in Table 3 will be considered as basic, without considering metrics describing discrepancies, as they do not satisfy axioms A3-A7. In fact, we will assume that all data used obey some distribution law, then the similarity metric in terms of statistical distance can be described as:

$$||P - Q||_{TS} = \sup_{\mathcal{A} \in \mathcal{F}} |P(\mathcal{A}) - Q(\mathcal{A})| \qquad (22)$$

Where P and Q are the probabilities of distribution of the studied and reference values.

The second metric is the Helling distance, determined by the equation:

$$H^2(P,Q) = \frac{1}{2}\int_{\mathcal{X}} \left(\sqrt{p} - \sqrt{q}\right)^2 d\lambda = 1 - \int_{\mathcal{X}} \sqrt{pq}\, d\lambda \qquad (23)$$

Where $P(dx) = p(x)\lambda(dx)$ and $Q(dx) = q(x)\lambda(dx)$, and $p$ and $q$ – Radon-Nicodim measures defined for the reference and new states, calculated using the equation:

$$p(s) = \int_S f(s)ds \qquad (24)$$

Where $f$ is the density function of states according to the existing measure s.

The third metric applicable for evaluating D(s,s0) is the Levy-Prokhorov metric, calculated using the equation:

$$\pi(\mu,\nu) = \inf\{\varepsilon > 0: \mu(A) \le \nu(A^\varepsilon) + \varepsilon \wedge \nu(A) \le \mu(A^\varepsilon) + \varepsilon, \forall A \in B(X)\} \qquad (25)$$

For probability measures, $\pi(\mu,\nu) \le 1$, where ε is the neighborhood of two probability measures μ and ν, on the entire space of Borel sets B(M), while ε is the neighborhood of a subset A defined as:

| | |
|---|---|
| $$A^{\varepsilon} = \{x: \exists y \in A, d(x,y) < \varepsilon\} = \bigcup_{p \in A} B_{\varepsilon}(p)$$ | (26) |

Where $B_{\varepsilon}(p)$ is a sphere of radius ε centered at p.

The fourth metric to consider is the Wasserstein metric:

| | |
|---|---|
| $$\mathcal{W}_p(\mu,\nu) = \left( \inf_{\gamma \in \Gamma(\mu,\nu)} \int_{M \times M} d(x,y)^p d\gamma(x,y) \right)^{\frac{1}{p}}$$ | (27) |

Where Γ(μ,ν) denotes the set of all measures on M×M with marginal (partial) distributions μ and ν for the first and second parameters, respectively; $p \geq 1$ is the moment number of the distribution, d(x,y) is the metric on X, and γ(x,y) is the measure on X×X.

The fifth metric considered in our work is the Mahalanobis distance. The basic meaning of this metric is described by the equation:

| | |
|---|---|
| $$d_M(x,y,\Sigma) = \sqrt{(x-y)^T \Sigma^{-1} (x-y)}$$ | (28) |

Where ∑ is the covariance matrix.

The last statistical metric we will consider is the integral probability metric, calculated using the equation:

| | |
|---|---|
| $$IPM_{\mathcal{F}}(P,Q) = \sup_{f \in \mathcal{F}} \left| \mathbb{E}_{X \sim P} f(X) - \mathbb{E}_{Y \sim Q} f(Y) \right| = \sup_{f \in \mathcal{F}} \left| Pf - Qf \right|$$ | (29) |

In addition to probabilistic metrics, there are metrics developed specifically for evaluating the transfer learning capabilities of ML models, for solving Domain Adaptation (DA) and Domain Generalization (DG) problems [83,84]. A list of these metrics is presented in Table 4.

**Table 4.** List of metrics used in Domain Adaptation (DA) and Domain Generalization (DG).

| Metric name | The main idea | Application in DA/DG |
|---|---|---|
| Maximum Mean Discrepancy (MMD) [85,86] | Criterion for statistical equalization of distributions | The main metric in Deep Adaptation Networks (DAN), Joint Adaptation Network (JAN), RTN |
| Wasserstein Distance (Earth Mover's Distance) [73] | The metric of "mass transfer" between distributions | Used for domains with different geometries, Wasserstein DA (WDAN) |
| CORAL (CORrelation ALignment) [87,88] | Alignment of second moments distributions | Deep CORAL — a simple and effective technique for evening out statistics in DA |
| Proxy A-distance (PAD) [89] | Quantitative assessment of domain differences | Used to assess the degree of difference between domains |
| H-divergence (Ben-David et al.) [79] | Theoretical criterion for distinguishing between domains | Analysis of the generalizing ability of models, DANN, ADDA |
| Jensen-Shannon Divergence [78] | Information-entropy measure of convergence of distributions | Used in GAN-based DA (e.g., CoGAN) for distribution alignment |
| Central Moment Discrepancy (CMD) [90] | Multicore alignment in deep networks | To improve the consistency of statistics from different moments in DA |
| Sliced Wasserstein Distance (SWD) [91] | Wasserstein distance for distribution projections | Used for high-dimensional representations (UNIT, SWD-based DA) |
| Projection Metric [92] | Methods for finding invariant subspaces | Helps find subspaces with common features between domains |
| Domain Discrepancy Error [93] | Model tolerance assessment measure h | Quantitative assessment of differences between domains |
| Gradient Similarity (GS) [94] | Regularization for gradient smoothing | Used, for example, in DeepCORAL for stable learning |
| Leave-One-Domain-Out Variance [95] | Assessment of model stability between domains | Assessment of model variability when excluding one of the domains |
| Domain-Agnostic Metric (DAM) [96] | Metric for DANN-DG type algorithms | Allows you to create models that are resistant to domain changes |
| Invariant Risk Minimization (IRM) [97] | Penalty for gradient variability for invariance | Building models with common invariant features across domains |

The main criterion for selecting the most suitable metric is that it satisfies all the requirements of the axioms and the analogy theorem in both predicate logic and Hoare logic. An analysis of the mathematical foundations of all the metrics presented showed that the Wasserstein metric satisfies the requirements of the analogy theorem and axioms, but this metric has high computational complexity. Despite this drawback, the Wasserstein metric will be used for further reasoning and calculations.

## 2.4. Definition of ε-neighborhood, δ, and γ based on Wasserstein metric

To determine the methods for calculating ε-neighborhood, δ, and γ, we will analyze the relationships between the Wasserstein metric and ε-neighborhood, δ, and γ. For this purpose, we will formulate the problem in the language of probability theory. Let there be a metric space *(X, d)* with metric d and a space of probability measures on X with finite p-moment, then the true (analytically unknown) distributions are given in two domains $P, Q \in \mathcal{P}_p(X)$ and empirical measures with independent samples $\{X_i\}_{i=1}^{n} \sim P$ and $\{Y_j\}_{j=1}^{m} \sim Q$ will be described as:

| | |
|---|---|
| $P_n = \frac{1}{n}\sum_{i=1}^{n} \delta_{X_i} \; u \; Q_m = \frac{1}{m}\sum_{j=1}^{m} \delta_{Y_j}$ | (30) |

Then the Wasserstein metric of order p will be described by the equation:

| | |
|---|---|
| $\mathcal{W}_p(P, Q) = \left( \inf_{\gamma \in \Gamma(P,Q)} \int_{X \times X} d(x, y)^p d\gamma(x, y) \right)^{\frac{1}{p}}$ | (31) |

Where Γ(P, Q) is the set of all joint distributions with margins P and Q. In this case, equation (31) will represent a distance measure reflecting the minimum cost of data transfer.

To apply the Wasserstein metric, we formalize the conditions of the analogy theorem, i.e., we define the properties *F, L: X → {0,1}*, and for predicates, model variables, and distances, we require that:

| | |
|---|---|
| $\forall x \in X, d(x, x_0) \leq \varepsilon \Rightarrow F(x) = L(x)$ | (32) |

That is, the analogy holds true. And:

| | |
|---|---|
| $\exists x \in X, d(x, x_0) > \varepsilon \Rightarrow F(x) \neq L(x)$ | (33) |

When the analogy breaks down. Where $d(x, x_0) = \mathcal{W}_p(\delta_x, \delta_{x_0})$.

Accordingly, it is necessary to estimate ε using empirical measures that can be calculated based on existing data.

One option for estimating the metric given by equation (31) is to estimate it using confidence intervals and empirical measures [98,99]. Let us consider this approach in more detail. Thus, let there be a constant C>0 such that for all n, the following holds:

| | |
|---|---|
| $\mathbb{E}W_p(P_n, P) \leq Cr_n(d, p), \; r_n(d, p) = \begin{cases} n^{-\frac{1}{2}}, & d < 2p \\ n^{-\frac{1}{2}} \log^{\frac{1}{2}} n, & d = 2p \\ n^{-\frac{1}{d}}, & d > 2p \end{cases}$ | (34) |

Where d is the actual dimension of space X (for example, the local dimension of measures).

If the amount of data is large enough and covers most of the values taken by the random variable under study (dense sampling), then the probability P can be described by a sufficiently smooth distribution, in which case for any t>0 we have:

| $\mathbb{P}\left(\left\|W_p(P_n,P)-\mathbb{E}W_p(P_n,P)\right\|>t\right)\leq 2\exp\left(-C_1 n t^{\alpha}\right)$ | (35) |
|---|---|

Where $C_1$ depends on the properties of the measure and X, and α depends on the geometry of space and the properties of the measure (for sub-Gaussian distributions, α=2, and in the case of compact support, Lipschitz cost, and the presence of isoperimetric/transport properties of the measure, P α=d/p).

Taking all the above into account, we can present rules for the practical assessment of all parameters of interest to us, namely, the practical assessment of $W_p(P,Q)$ will be as follows:

| $W_p(P_n,Q_m)\approx W_p(P,Q)\pm\epsilon$ | (36) |
|---|---|

Where $\epsilon\leq\epsilon_{stat}$ and $\epsilon_{stat}$ is determined by the equation:

| $\epsilon_{stat}\lesssim C\left(r_n(d,p)+r_m(d,p)\right)$ | (37) |
|---|---|

Then the confidence interval for Monte Carlo samples will be as follows:

| $\varepsilon=W_p(P_n,Q_m)+z_{1-\alpha}\sigma_W$ | (38) |
|---|---|

Where $\sigma_W$ is the standard deviation of Monte Carlo samples; $z_{1-\alpha}$ is the quantile of the normal distribution; *1-α* is the confidence level. In a more general representation, not tied to the distribution law, the confidence interval estimate is constructed as follows:

| $\varepsilon=Q_{1-\alpha}\left(\left\{W_p^{(b)}\right\}_{b=1}^{B}\right)$ | (39) |
|---|---|

Where $Q_{1-\alpha}$ is the quantile of the Monte Carlo estimation distribution $W_p^{(b)}$ corresponding to the confidence level 1-α.

Equations (38) and (39) fully satisfy the requirements of axiom M3.1 for sufficiently large n and m.

The estimation of the transformation stability parameter is based on equations (30) and (31), namely, let there be a certain transformation that models the change in state (e.g., a program or transformer) such that $\varphi: X\rightarrow X$, then the maximum deviation of the point distribution will be described by the equation:

| $\gamma=\sup_{x\in X} W_p\left(\delta_x,\delta_{\varphi(x)}\right)=\sup_{x\in X} d\left(x,\varphi(x)\right)$ | (40) |
|---|---|

At the same time, if the transformation φ is L-Lipschitz, then for any P and Q we have:

| $W_p(\varphi_\# P,\varphi_\# Q)\leq L\cdot W_p(P,Q)$ | (41) |
|---|---|

Where the symbol # denotes a measure defined through the transformation φ.

If a data set (sample) is available, the stability parameter of the transformation can be estimated as:

| $\gamma\approx Q_{1-\beta}\left(\{d(x,\varphi(x))\}_{x\in\text{выборке}}\right)$ | (42) |
|---|---|

Where $Q_{1-\beta}$ is the quantile of the distribution of metric d. In the case of β=0.05, we obtain a quantile corresponding to a 95% confidence probability.

Now let us evaluate the local parameter δ. So, let the space be divided into classes with probability measures $\mu_i \in \mathcal{P}_p(X)$, where i takes values from 1 to K, then the distance between classes will be calculated as:

$$D_{ij} = W_p(\mu_i, \mu_j), i \neq j \quad (43)$$

Accordingly, under the conditions of the analogy theorem, the parameter δ must be chosen in such a way as to guarantee the reliability of the local equivalence of the property. This result can be achieved if δ is calculated using the equation:

$$\delta = \frac{1}{2} \min_{i \neq j} (D_{ij} - r_i - r_j) \quad (44)$$

Where $r_i$ is the radius of the class, determined by the equation:

$$r_i = \sup_{x \in \text{класс } i} W_p(\delta_x, \mu_i) \quad (45)$$

Representing the parameter δ in the form of equation (44) ensures that points within the vicinity of radius δ belong to the same class, and in this case, the minimum distance between classes will be maintained.

In conclusion, we will analyze the conditions for the correctness of the analogy for the final selection of parameters. As shown in part 2, the analogy is valid if $\gamma < \min\left(\varepsilon, \frac{\delta}{2}\right)$, i.e., if the shift is less than the threshold of the error in estimating the distance between domains and half of the local threshold for properties, then the analogy is valid; otherwise, the analogy will be violated. To ensure the reliability of the condition for the existence of analogy, the safety parameters *μ, ξ>0*, then the condition for the analogy will be as follows:

$$\gamma < \min\left(\varepsilon - \eta, \frac{\delta - \xi}{2}\right) \quad (46)$$

Then, to ensure consistency with equations (38)-(45), we have:

$$\eta = z_{1-\alpha} \sigma_W \quad (47)$$

And

$$\xi = \mathcal{Q}_{0.99}\left(d\big(x, \varphi(x)\big)\right) - \mathcal{Q}_{0.95}\left(d\big(x, \varphi(x)\big)\right) \quad (48)$$

Having determined the conditions for guaranteed fulfillment of the analogy, we can move on to probabilistic proof. Let the probability of non-fulfillment of the analogy be estimated as:

$$\mathbb{P}\left(\gamma \geq \min\left(\varepsilon - \eta, \frac{\delta - \xi}{2}\right)\right) \leq \alpha + 2\beta \quad (49)$$

Then, we get:

$$\mathbb{P}\left(\gamma \geq \min\left(\varepsilon - \eta, \frac{\delta - \xi}{2}\right)\right) \leq \mathbb{P}(\gamma \geq \varepsilon - \eta) + \mathbb{P}\left(\gamma \geq \frac{\delta}{2} - \xi\right) \quad (50)$$

Applying concentration inequalities to inequality (50), each of the terms on the right-hand side is less than or equal to β, we obtain that the probability of violating the analogy is:

| $\mathbb{P}(violations) \leq 2\beta$ | (51) |
|---|---|

Summarizing all the above considerations, we obtain the following probabilistic estimates of the boundaries of applicability of analogy based on the Wasserstein metric:

| Parameter | Equation | Recommendations |
|---|---|---|
| $\varepsilon$ | $W_p(P_n, Q_n) + z_{1-\frac{\alpha}{2}}\sigma_W$ | $\alpha = 0.05, B \geq 1000$ |
| $\gamma$ | $Q_{1-\beta}\left(d\big(x, \varphi(x)\big)\right)$ | $\beta = 0.05$ |
| $\delta$ | $\delta = \frac{1}{2}\min_{i \neq j}(D_{ij} - r_i - r_j)$ | --- |
| Condition | $\gamma < \min\left(\varepsilon - \eta, \frac{\delta}{2} - \xi\right)$ | $\eta = z_{0.975}\sigma_W$<br>$\xi = \mathcal{Q}_{0.99} - \mathcal{Q}_{0.95}$ |

Thus, a complete structure for evaluating the parameters of the analogy theorem based on a probabilistic approach is provided. The application of the Monte Carlo method allows for preliminary evaluation and comparison of parameters. Let us consider in more detail the application of the theorem in the field of statistical modeling and machine learning.

## 3. Results

### 3.1. Verification of the analogy theorem on model data

In the final part of this paper, we will consider an example of applying the analogy theorem, implemented in the R programming language. Figure 1 shows a block diagram of the algorithm for applying the theorem with the Wasserstein metric to compare artificially generated data.

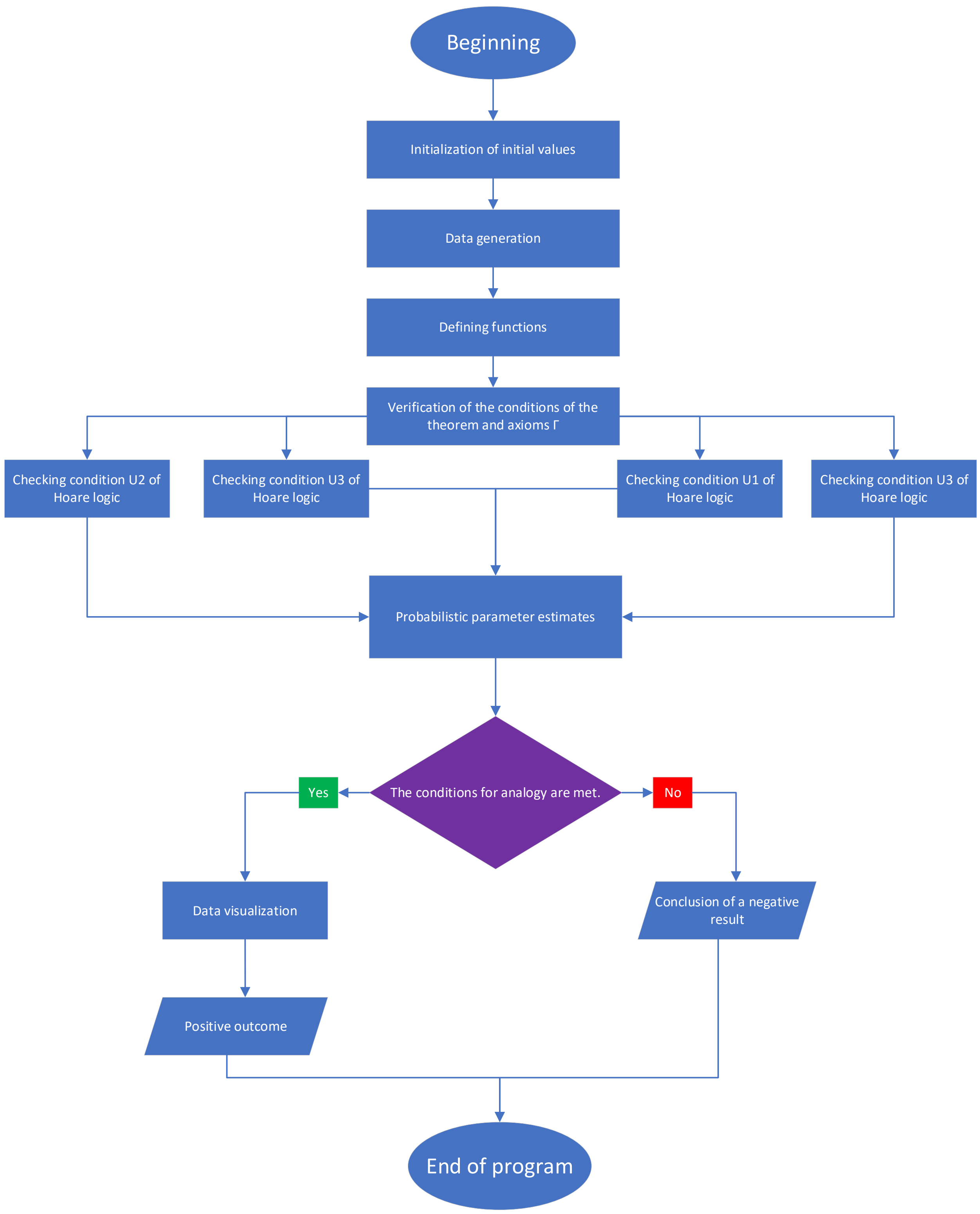

**Figure 1** – Block diagram of the data analysis algorithm in accordance with the analogy theorem.

Appendix A presents the program code in the R programming language that implements this algorithm. Table 5 presents the main input parameters of the algorithm, and Figure 2 presents the results of comparing two model types of scattering.

**Table 5.** Input data for the algorithm for checking data for compliance with the analogy theorem.

| Parameter | Meaning option 1 | Parameter assignment |
|---|---|---|
| n | 500 | Number of attribute values |
| d | 2 | Number of variables |
| P_sample | See Supplementary script ChekTeoremNNRF.R | Multivariate distribution Random distribution of numbers obeying the normal distribution law |
| Q_sample | See Supplementary script ChekTeoremNNRF.R | P_sample |
| epsilon_val | 0.5 | ε- neighborhood of a theorem |
| delta_val | 2.0 | δ- neighborhood of a theorem |
| gamma_val | 0.05 | γ- neighborhood of a theorem |
| subset_size | 100 | Sample size from a multivariate normally distributed sample |
| class1 | See Supplementary script ChekTeoremNNRF.R | Multivariate distribution Random distribution of numbers obeying the normal distribution law |
| class2 | See Supplementary script ChekTeoremNNRF.R | Multivariate distribution of values obeying the normal distribution law |
| eta | Eq. (31) | Reliability correction |
| xi | Eq. (32) | Reliability adjustment |
| Result of theorem verification | Verified | |

Figure 2 shows the results of applying the parameters of the analogy theorem to the model data.

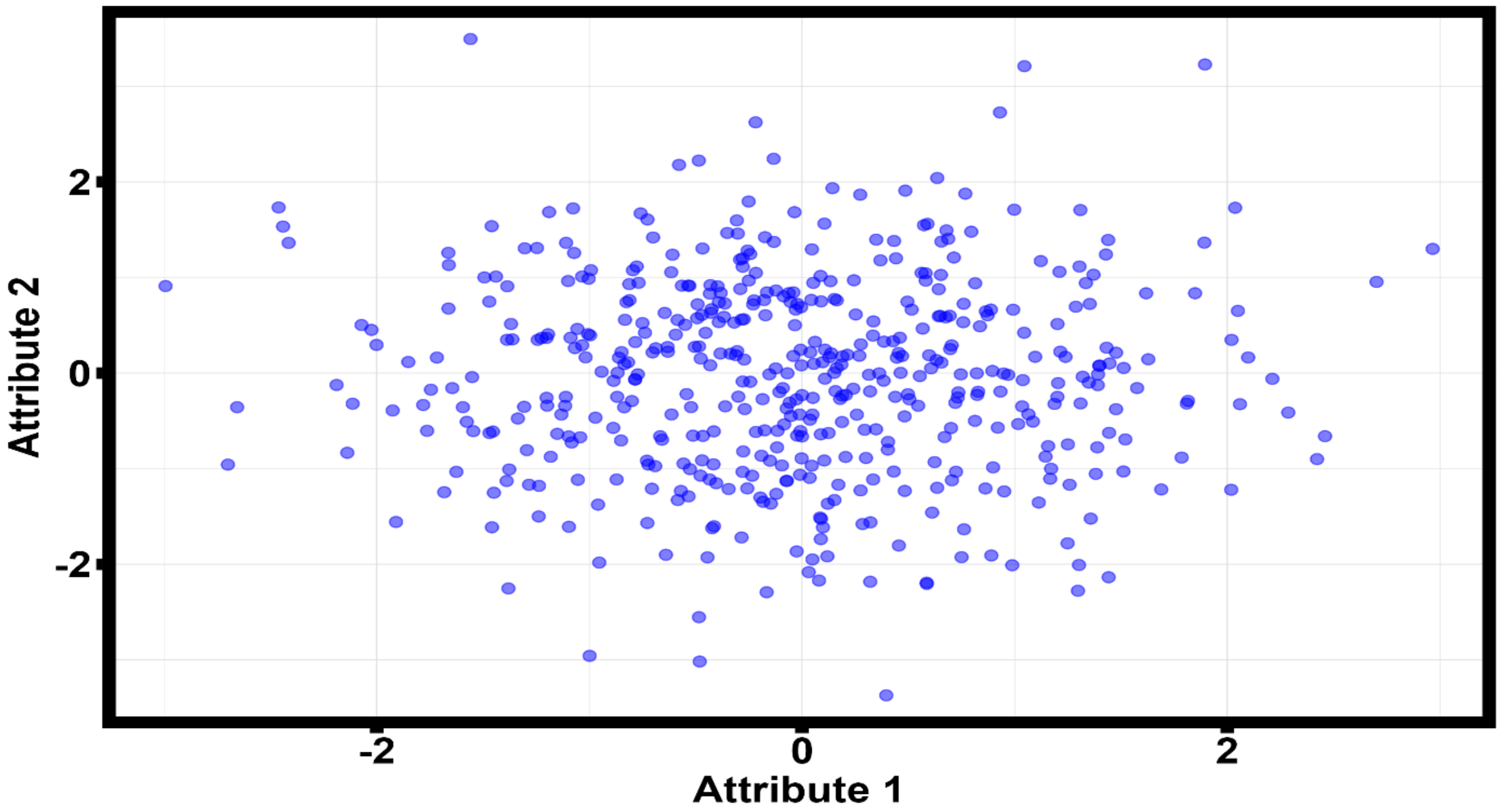

**Figure 2** – The distribution of data in the source and target domains is completely identical.

As a result of calculations, the Wasserstein metric is close to zero. Table 6 shows the estimation of the Wasserstein metric using the bootstrap method (the number of sample repetitions is 1000).

**Table 6.** Results of estimating the Wasserstein metric using the bootstrap method.

| Average metric value | Sd | Lower limit of the 95% interval | Upper limit of the 95% interval |
|---|---|---|---|
| 0.080 | 0.021 | 0.052 | 0.128 |

The results obtained from verifying the analogy theorem on model data show that for there to be an analogy between domains, the data must have a metric value close to zero, i.e., they must be practically indistinguishable from each other. It should be noted that during the simulation, we were guided by the normal distribution law of data; however, in practice, the distribution of data does not always correspond to the normal law, and before checking the correspondence between domains, it is necessary to evaluate the distribution law. One of the methods is presented in our applied works [100-102]. One of the main problems with applying the Wasserstein metric to big data is the inability to scale it, but when working with large data sets, Sliced Wasserstein Distance [91] can be used to approximate real data using the Wasserstein metric. Table 7 presents the results of the analysis of computation time on the same model data (Supplementary script ChekTeoremNNRF.R) under the condition that all conditions of the analogy theorem are satisfied, conducted using a direct evaluation of the Wasserstein metric and using Sliced Wasserstein Distance.

**Table 7.** Comparison of the execution times of two metrics for assessing the existence of Wasserstein analogy and Sliced Wasserstein Distance.

| Metric type | Calculation time, s | Number of projections | Are the conditions of the theorem satisfied? |
|---|---|---|---|
| Wasserstein | 2.53 | 1000 | The theorem has been verified. |
| Sliced Wasserstein Distance | 2.24 | 1000 | The theorem has been verified. |
| Wasserstein | 5.01 | 2000 | The theorem has been verified. |
| Sliced Wasserstein Distance | 4.31 | 2000 | The theorem has been verified. |

The results of comparing the execution times of the two methods show that as the number of projections increases, the calculation time using the Sliced Wasserstein Distance metric differs more significantly from the calculation time using the Wasserstein metric. Accordingly, for small domain sizes (up to 1000), both metrics can be used without a significant gain in calculation speed, while for larger domain sizes (over 1000 values), it is better to use the Sliced Wasserstein Distance metric to speed up calculations. At the same time, there is no violation of the strictness of the analogy theorem.

Now let's consider the application of analogy to machine learning tasks to standard Domain Adaptation (DA) and Domain tasks.

### 3.2. Knowledge transfer experiment

To verify the application of the analogy theorem in knowledge transfer problems, we selected several standard problems in Domain Adaptation (DA) and Domain. For the experiment, we used two types of publicly available datasets: MNIST [103], containing approximately 1,280,000 training data points of handwritten digits and approximately 512,000 in the test dataset. The images are presented in grayscale, measuring 28x28 pixels and containing 784 features per image.

The second dataset used to test the model and perform domain adaptation is the USPS [104] dataset, containing 512,000 training data points and 512,000 test data points. The images of handwritten digits are presented in grayscale with a resolution of 16x16 pixels and contain 256 features per image.

When training the models, the original MNIST images were compressed to 16x16. When building a multi-class classifier, two machine learning models were used: a multilayer convolutional neural network [105,106] and a random forest [107,108], the general view of which is shown in Figure 3.

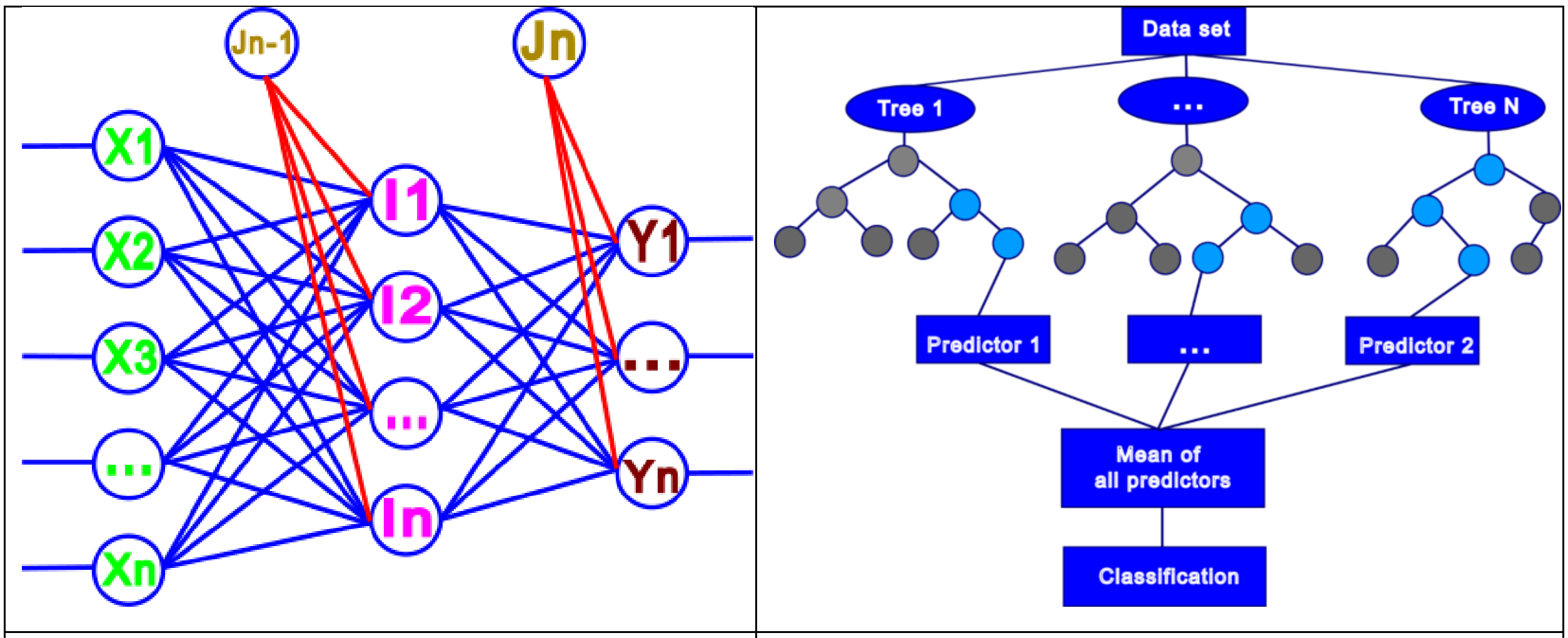

a) b)

**Figure 3** – General view of neural networks used to verify the analogy theorem on MNIS datasets. a) Convolutional neural network; b) Random Forest model. Where X1, X2, X3,…, Xn are explanatory parameters; Y1..Yn are output parameters; I1,I2,…,In and K1, K2,…,Kn are hidden parameters; Jn-1, Jn are correction parameters.

There were 16 explanatory variables and 10 explained variables. The convolutional neural network had two hidden layers with 64 hidden parameters in the first layer and 32 in the second layer, respectively. There were 500 trees in the random forest model.

The choice of this convolutional neural network architecture was dictated by the balance between accuracy and speed of computation [101]. When testing the architecture, the logistic, softplus, and hyperbolic tangent functions, defined by the equations:

| | |
|---|---|
| $f(x) = \frac{1}{1 + \exp(-x)}$ | (52) |
| $f(x) = \ln(1 + \exp(x))$ | (53) |
| $f(x) = \tanh(x)$ | (54) |

The highest classification accuracy in the convolutional neural network model was achieved using the activation function described by equation (52). The decision threshold remained constant at 0.5 and did not change during the study. For the random forest model, the decision threshold was taken to be equal to the threshold of the convolutional neural network.

Table 8 shows the main quality metrics for multi-class classification of machine learning models trained and tested on MNIST data without using data from the USPS dataset.

**Table 8.** Quality metrics for the random forest and convolutional neural network models trained and tested on MNIST data.

| Metric | Random Forest | CNN |
|---|---|---|
| Accuracy [109] | 0.896 | 0.844 |
| Sensitivity [109] | 0.895 | 0.841 |
| Specificity [109] | 0.988 | 0.983 |
| Precision [109] | 0.897 | 0.842 |
| Cohen's Kappa [110,111] | 0.884 | 0.826 |
| Matthews correlation coefficient [109] | 0.884 | 0.826 |
| F1-score [112] | 0.895 | 0.841 |

The analysis of quality metrics (Table 8) shows that the random forest model has higher metrics compared to the convolutional neural network model. Table 9 presents the quality metrics of models for classifying USPS data without using the Domain Adaptation (DA) method.

**Table 9.** Quality metrics of machine learning models on USPS data

| Metric | Random Forest | CNN |
|---|---|---|
| Accuracy [109] | 0.061 | 0.092 |
| Sensitivity [109] | 0.068 | 0.098 |
| Specificity [109] | 0.895 | 0.898 |
| Precision [109] | 0.077 | 0.105 |
| Cohen's Kappa [110,111] | -0.051 | -0.015 |
| Matthews correlation coefficient [109] | -0.049 | -0.015 |
| F1-score [112] | 0.070 | 0.096 |

The model quality metrics obtained from the USPS data show that the models do not classify the data but guess it randomly. To improve the model quality metrics, the Domain Adaptation (DA) method was applied without using the analogy theorem and using the Wasserstein Distance metric. Figure 4 shows the scatter of the first principal components [113] in the MNIST and USPS datasets without adaptation, and the density of the Wasserstein distances and the scatter of the data after adaptation.

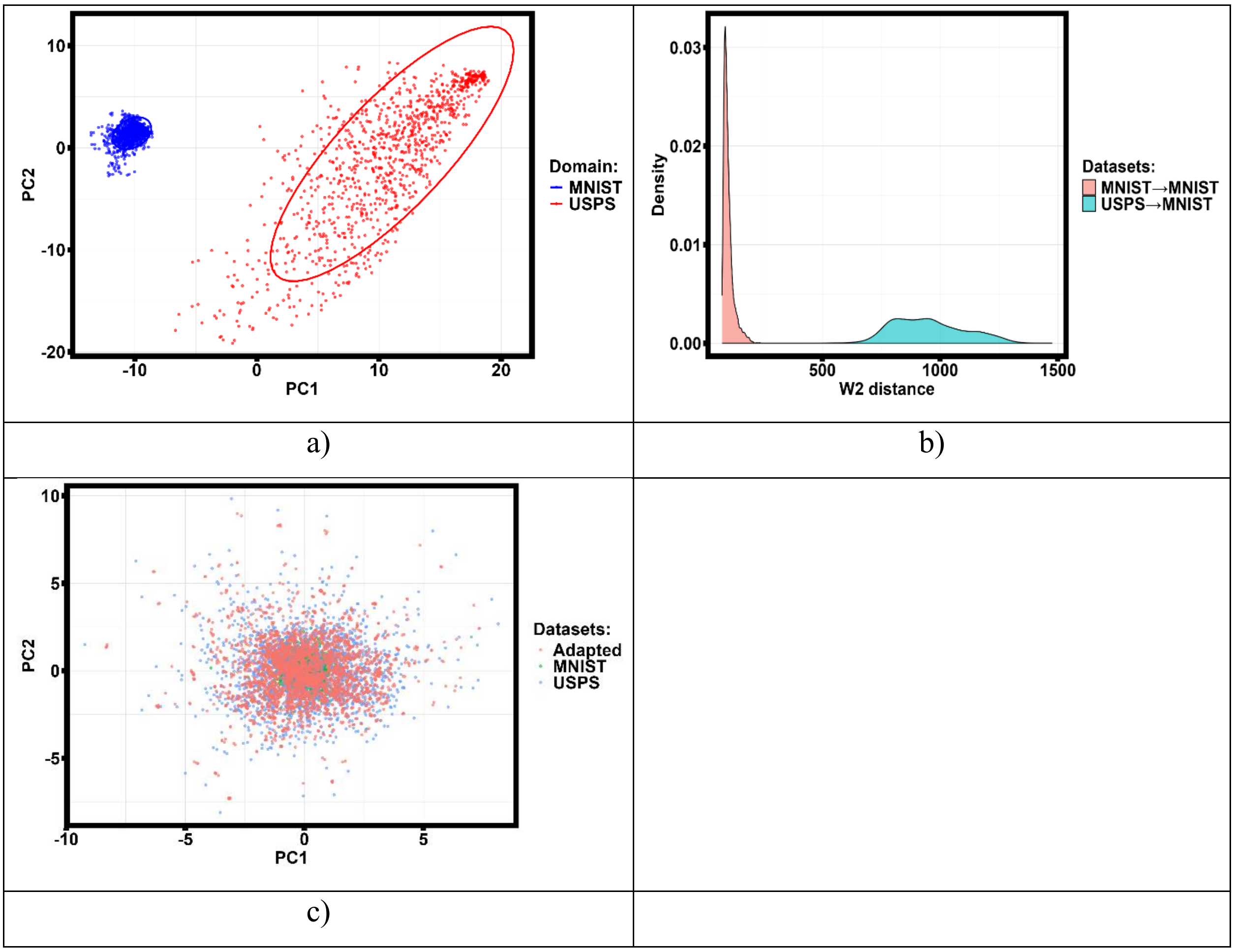

**Figure 4** – a) scatter plot of the first principal components of the original data (without adaptation), b) Wasserstein distance distribution density for the MNIST and USPS datasets, and c) scatter plot of the first principal components after data adaptation.

The analysis shows that the first principal components of the original data have significant differences (Figure 4a), which is also confirmed by the Wasserstein distance distribution (Figure 4b). Bringing the data to a single scale while preserving the Wasserstein distance distribution (Figure 4c) allows us to form an adapted dataset and obtain more universal machine learning models. Table 10 presents the quality metrics of machine learning models after applying the Domain Adaptation (DA) method with the Wasserstein Distance metric.

**Table 10.** Quality metrics of machine learning models of adapted data using the Wasserstein Distance metric without applying the analogy theorem.

| Metric | Random Forest | CNN |
|---|---|---|
| Accuracy [109] | 0.892 | 0.853 |
| Sensitivity [109] | 0.876 | 0.839 |
| Specificity [109] | 0.986 | 0.982 |
| Precision [109] | 0.897 | 0.838 |
| Cohen's Kappa [110,111] | 0.877 | 0.833 |
| Matthews correlation coefficient [109] | 0.799 | 0.759 |
| F1-score [112] | 0.883 | 0.838 |

Comparison of machine learning model quality metrics after applying Domain Adaptation technology (DA) (Table 10) with metrics before applying the technology (Table 9) and metrics obtained on MNIST data (Table 8) show that the model metrics are comparable to the results obtained on models trained and tested only on MNIST data.

Figure 5 shows the dependence of the accuracy of CNN and RF machine learning models on the size ε of the neighborhood of the analogy theorem.

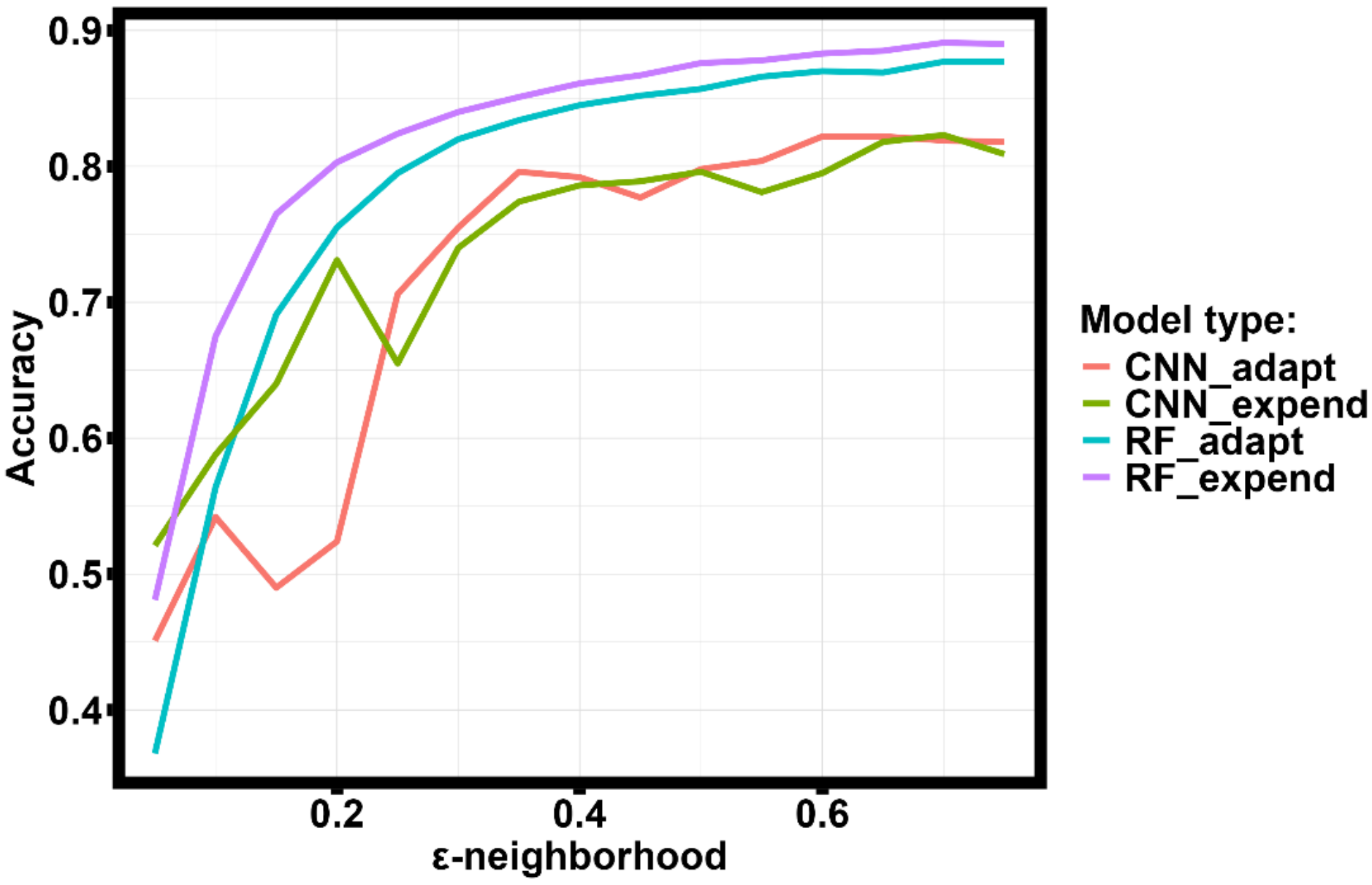

**Figure 5** – Dependence of the accuracy of models with different types of data correction on the size of the ε-neighborhood. Adapt – adapted dataset; Expend – dataset expanded by the ε-neighborhood.

Analysis of the obtained dependence shows that machine learning models achieve the accuracy of models obtained using the Wasserstein Distance metric with an ε neighborhood size of 70%. The random forest model demonstrates higher metrics on the expanded dataset compared to the adapted dataset for all investigated values of ε neighborhood.

In all cases considered, the parameters of the analogy theorem were checked, and the calculation results can be verified using the scripts provided in the appendix to the publication.

## 4. Discussion of results

A study of the proposed analogy theorem shows that the Domain Adaptation problem can be solved by integrating Wasserstein distance and contrastive learning. This solution eliminates the need to apply the λ-distance between the source and target domains, as proposed in [114]. If we consider this problem in more detail, the authors [114], based on the theorem about the impossibility of adapting a model to new data without the condition of distribution consistency, describe the error boundaries as:

$$\mathcal{E}_{\mathbb{Q}} \leq \lambda + \mathcal{E}_{\mathbb{P}}(h) + \frac{1}{2} d_{\mathcal{H}\Delta H}(\mathbb{Q}, \mathbb{P}) \quad (55)$$

Where λ is the error of "perfect hypothesis compatibility," which critically depends on the existence of hypotheses that work effectively in both domains. Accordingly, without this term, adaptation becomes impossible. In practice, this condition is often not met (for example, when distributions shift in medical images or texts).

An alternative solution to the problem of minimizing the distance between distributions was proposed in [115]. This solution is based on covariant shift, i.e., on the coincidence of conditional distributions for the domains being compared. The second condition [115] is the possibility of approximating the shift of distributions through mass transport. In both cases ([114] with λ and [115]), sample weighting is required, which, as we show, degrades the generalization ability when using Wasserstein DA due to a conflict with contrastive learning objectives. Meanwhile, the conditions of the analogy theorem do not require sample weighting across the entire data set. Furthermore, the computational complexity of calculating the Wasserstein distance is $O(n^3)$, which makes it difficult to apply to large datasets; in our work, we solve this problem by switching to the Sliced Wasserstein distance (SWD) metric when working with large datasets, which reduces the computational complexity to O(n log n). In our work, we use two approaches based on metrics: Wasserstein distance and Sliced Wasserstein distance (SWD), as they fully satisfy the axioms and conditions of the analogy theorem and allow us to work effectively with data sets of various sizes.

A comparison of the proposed analogy theorem with the approaches demonstrated in [114, 115] to the problem of domain adaptation in terms of generalization ability and scalability shows that aligning only the source domains without adapting to the target does not guarantee generalization [116]. The proposed analogy theorem solves this problem by forcibly separating clusters corresponding to different classes in the feature space, which is implemented through a contrastive learning mechanism. In addition, the presented concept of dynamic balls in Wasserstein space (see [117] for details) allows the radius ε to be adapted based on the local data density [117]. A comparison of the proposed approach with Domain-Adversarial Neural Networks (DANN) [118] reveals the peculiarities of applying the analogy theorem: the use of Wasserstein and Sliced Wasserstein metrics allows working with local minimization of Wp rather than with divergences, which helps to avoid the so-called “mode collapse.”

The experimental cases considered in our work show that the application of the analogy theorem in the adaptation of MNIST → USPS leads in all cases to an increase in classification accuracy, both on the convolutional neural network model and on the random forest model. The increase in classification accuracy is related to both the level of noise in the data and the data transformation. Our results show that the level of noise has a more significant impact on the increase in classification quality than data compression. When using the approach used in [116], classification accuracy decreases significantly, which is not observed in our work (except in cases of significant shifts in the target domain data).

Despite all its positive aspects, the analogy theorem has several features that require further study and refinement. In this work, we did not analyze sensitivity to the choice of $x_0$ (reference element), which is quite important in such areas of application as medicine or materials science. In addition, the approach implemented in [116] describes the formal error boundaries of the target domain, whereas this issue is not considered in the present work. This feature is also important for eliminating vulnerability to the so-called “adaptation gap” that manifests itself in medical applications. The application of the Wasserstein metric, as in the case of [117], requires the selection of p and Wp, which in turn leads to instability in semantic segmentation tasks [119]. These problems can be eliminated by automating the selection of representative points using few-shot learning [120] and extending the application of the theorem to new classes in the target domain. The problem of the stability of the Wasserstein metric in semantic segmentation tasks can also be solved by introducing target domain entropy estimates. We plan to address the problems identified in the analogy theorem in our future work.

## 5. Conclusion

A summary of the research results shows that the analogy theorem removes the fundamental limitations of previous work in the field of domain adaptation. Moving away from the λ parameter allows for greater versatility of the method for domains with inconsistent data distribution laws. The combination of Wp with contrastive learning allows the semantic structure of the data to be preserved. The parameters of the analogy theorem ε, δ, and γ allow for engineering-interpretable guarantees of the model's applicability, making it applicable to most practical tasks.

Verification of the analogy theorem on data generated by the Monte Carlo method and on simple images contained in the MNIST and USPS datasets shows that applying the analogy theorem to real cases allows us to obtain high metrics for the quality of multi-class data classification and to obtain more universal machine learning models.

Despite all the positive aspects of the presented analogy theorem, there are several problems that can be solved using already developed technologies and approaches which will be done in our future work.

**Acknowledgments:**
The author would like to thank Andrey Stepaskin, Senior Researcher at the Center for Composite Materials at NRTU MISIS, for his financial support in conducting the research.
**Funding source**:

The work was supported by the Russian Science Foundation, grant 23-73-00131.

**Author Contributions:**

All authors have read and agreed to the published version of the manuscript.

**Conflicts of Interest:** The authors declare no conflict of interest.

**Use of AI:** Three artificial intelligence systems, ChatGPT-5, DeepSek 3.1, and Perplexity, were used to conduct preliminary reviews of the publication.